\documentclass[sigconf]{acmart}

\AtBeginDocument{%
  \providecommand\BibTeX{{%
    \normalfont B\kern-0.5em{\scshape i\kern-0.25em b}\kern-0.8em\TeX}}}

\copyrightyear{2022}
\acmYear{2022}
\setcopyright{acmlicensed}\acmConference[GECCO '22 Companion]{Genetic and Evolutionary Computation Conference Companion}{July 9--13, 2022}{Boston, MA, USA}
\acmBooktitle{Genetic and Evolutionary Computation Conference Companion (GECCO '22 Companion), July 9--13, 2022, Boston, MA, USA}
\acmPrice{15.00}
\acmDOI{10.1145/3520304.3534012}
\acmISBN{978-1-4503-9268-6/22/07}

\usepackage{physics}
\usepackage{qcircuit}
\usepackage{tikz}
\usepackage{etoolbox}
\usepackage{bm}
\usepackage{bbm}

\newcommand{\eg}{{\em e.g.}}

\begin{document}

\title{Evolutionary Quantum Architecture Search for\\
Parametrized Quantum Circuits}


\author{Li Ding}
\affiliation{%
  \institution{University of Massachusetts Amherst}
  \country{}
}
\email{liding@umass.edu}

\author{Lee Spector}
\affiliation{%
  \institution{Amherst College}
  \institution{University of Massachusetts Amherst}
  \country{}
}
\email{lspector@amherst.edu}


\begin{abstract}
  Recent advancements in quantum computing have shown promising computational advantages in many problem areas. As one of those areas with increasing attention, hybrid quantum-classical machine learning systems have demonstrated the capability to solve various data-driven learning tasks. Recent works show that parameterized quantum circuits (PQCs) can be used to solve challenging reinforcement learning (RL) tasks with provable learning advantages. While existing works yield potentials of PQC-based methods, the design choices of PQC architectures and their influences on the learning tasks are generally underexplored. In this work, we introduce EQAS-PQC, an evolutionary quantum architecture search framework for PQC-based models, which uses a population-based genetic algorithm to evolve PQC architectures by exploring the search space of quantum operations. Experimental results show that our method can significantly improve the performance of hybrid quantum-classical models in solving benchmark reinforcement problems. We also model the probability distributions of quantum operations in top-performing architectures to identify essential design choices that are critical to the performance.
\end{abstract}



\maketitle

\section{Introduction}

Near-term quantum computing technologies will hopefully allow quantum computing
systems to reliably solve tasks that are beyond the capabilities of classical
systems~\citep{preskill2018quantum}. One of the most promising applications of
quantum computing is hybrid quantum-classical learning systems, which utilize
parameterized quantum operations and classical optimization algorithms to solve
the learning tasks. Prior works have demonstrated that parametrized quantum
circuits (PQCs)~\citep{benedetti2019parameterized} are able to handle a variety
of supervised and unsupervised tasks such as
classification~\citep{schuld2020circuit,havlivcek2019supervised,schuld2019quantum,li2021vsql})
and generative
modeling\citep{liu2018differentiable,zhu2019training,larose2020robust,huang2021experimental},
as well as provide proofs of their learning advantages~\citep{du2020expressive,jerbi2021parametrized}.
Some recent work~\citep{jerbi2021parametrized,skolik2021quantum} further shows
that PQCs can be used to construct quantum policies to solve the more complex
reinforcement learning problems, with an empirical learning advantage over
standard deep neural networks (DNNs). 

One of the key aspects behind the success of PQC-based algorithms is the
architectural designs of hybrid quantum learning frameworks. While prior
works~\citep{jerbi2021parametrized,skolik2021quantum,perez2020data} have either
identified some essential components, or empirically analyzed the influence of
different design choices on the learning performances, the development of
high-performance architectures of hybrid quantum systems nevertheless relies on
human ingenuity. On the other hand, architecture search methods, which aim to
automate the process of discovering and evaluating the architecture of complex
systems, have been extensively explored in classical learning systems, e.g.,
neural architecture search (NAS)~\citep{elsken2019neural}. More specifically,
recent works~\citep{lu2019nsga, ding2021optimizing} on combining genetic
algorithms with gradient-based optimization have demonstrated superior
performance in NAS and more generally optimizing deep neural networks. In the
context of quantum computing, common architecture search approaches, such as
greedy algorithms~\citep{grimsley2019adaptive,tang2021qubit}, evolutionary
algorithms~\citep{las2016genetic,du2020quantum,lu2021markovian}, reinforcement
learning~\citep{niu2019universal,kuo2021quantum}, and gradient-based
learning~\citep{zhang2020differentiable} have also been attempted to solve tasks
such as quantum control, variational quantum eigensolver, and quantum error
mitigation. However, most of these approaches target on optimizing either
specific pure quantum circuits or single-qubit quantum operations, instead of
more complex multi-qubit hybrid systems. Overall, automated search and
optimization of architectures of hybrid quantum learning systems have not been
sufficiently explored yet.

In this work, we aim to explore using genetic algorithms to automatically design
the architecture of hybrid quantum-classical systems that can solve complex RL
problems. We propose EQAS-PQC, an Evolutionary Quantum Architecture Search
framework for constructing complicated quantum circuits based on some
fundamental PQC circuits. We adopt the ideas of successful approaches in NAS
using genetic algorithms, which have more flexible architecture search spaces
and require less prior knowledge about architecture design. In our experiments,
we consider the benchmark RL environments from OpenAI Gym, which has been widely
used for RL research. Experimental results show that agents trained by using our
method significantly outperform the ones from prior work. We further analyze the
top-performing PQC architectures found by our method to identify the common
patterns that appear during the search process, which may provide insights for
the future development of hybrid quantum-classical learning systems.

\section{Preliminaries and Related Work}

In this section, we introduce the basic concepts of quantum computation related
to this work, and give a detailed description of parametrized quantum circuits
and their applications. 

\subsection{Quantum Computation Basics}

An $n$-qubit quantum system is generally represented by a complex Hilbert space
of $2^n$ dimensions. Under the bra-ket notation, the quantum state of the system
is denoted as a vector $\ket{\psi}$, which has unit norm
$\braket*{\psi}{\psi}=1$, where $\bra{\psi}$ is the conjugate transpose and
$\braket*{\psi}{\psi}$ represents the inner-product. The computation basis
states are represented as the tensor products of single-qubit computational
basis states, \eg, the two-qubit state $\ket{01}=\ket{0}\otimes\ket{1}$ where
$\ket{0}=\begin{bmatrix} 1\\0 \end{bmatrix}$ and $\ket{1}=\begin{bmatrix} 0\\1
\end{bmatrix}$.

A unitary operator $U$ acting on qubits is called a quantum gate. Some common
quantum gates are frequently used in this work, namely the single-qubit Pauli
gates Pauli-$X$, Pauli-$Y$, Pauli-$Z$ and their associated rotation operators
$R_x$, $R_y$, $R_z$. The matrix representations of Pauli gates are
\begin{equation}
  X = \begin{bmatrix}
    0 & 1 \\ 1 & 0
  \end{bmatrix},
  Y = \begin{bmatrix}
    0 & -i \\ i & 0
  \end{bmatrix},
  Z = \begin{bmatrix}
    1 & 0 \\ 0 & -1
  \end{bmatrix}.
\end{equation}
Given a rotation angle $\theta$, the matrix representations of rotation
operators are
\begin{align}
  \notag R_x(\theta) &= \begin{bmatrix}
    \cos\frac{\theta}{2} & -i\sin\frac{\theta}{2} \\ 
    -i\sin\frac{\theta}{2} & \cos\frac{\theta}{2}
  \end{bmatrix},\\
  \notag R_y(\theta) &= \begin{bmatrix}
    \cos\frac{\theta}{2} & -\sin\frac{\theta}{2} \\ 
    \sin\frac{\theta}{2} & \cos\frac{\theta}{2}
  \end{bmatrix},\\
  R_z(\theta) &= \begin{bmatrix}
    e^{-i\frac{\theta}{2}} & 0 \\ 0 & e^{i\frac{\theta}{2}}
  \end{bmatrix}.
\end{align}

If a quantum state of a composite system can not be written as a product of the
states of its components, we call it an entangled state. An entanglement can be
created by applying controlled-Pauli-Z gates to the input qubits.

A projective measurement of quantum states is described by an observable, $M$,
which is a Hermitian operator on the state space of the quantum system being
observed. The observable has a spectral decomposition
\begin{equation}
  M = \sum_m mP_m,
\end{equation}
where $P_m$ is the projector onto the eigenspace of M with eigenvalue $m$. Upon
measuring the state $\ket{\psi}$, the probability of getting result $m$ is given
by
\begin{equation}
  p(m) = \bra{\psi}P_m\ket{\psi},
\end{equation}
and the expectation value of the measurement is
\begin{equation}
  E(M) = \sum_m m\cdot p(m) = \bra{\psi}M\ket{\psi}.
\end{equation}

For a more detailed introduction to basic concepts of quantum computation, we
refer the readers to \citet{nielsen2002quantum}. 

\subsection{Parametrized Quantum Circuits}

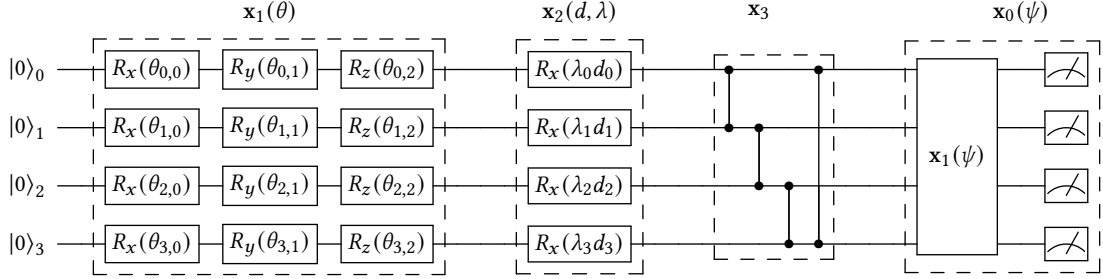
\begin{figure*}[t]
  \makebox[\linewidth]{\hspace{3em}
  \Qcircuit @C=1em @R=.8em {
  & & & \mathbf{x}_1(\theta) & & & & & \mathbf{x}_2(d,\lambda) &&&&& \mathbf{x}_3 &&&&&&& \mathbf{x}_0(\psi) \\
  &\\
  \lstick{\ket{0}_0} &\qw & \gate{R_x(\theta_{0,0})} & \gate{R_y(\theta_{0,1})} & \gate{R_z(\theta_{0,2})} &\qw &\qw &\qw &\gate{R_x(\lambda_{0}d_0)} &\qw &\qw &\qw &\ctrl{1} &\qw &\qw &\ctrl{3} &\qw &\qw &\qw &\multigate{3}{\mathbf{x}_1(\psi)} &\qw & \meter \gategroup{3}{3}{6}{5}{1.0em}{--}   \gategroup{3}{9}{6}{9}{1.0em}{--}  \gategroup{3}{13}{6}{16}{1.0em}{--}  \gategroup{3}{20}{6}{22}{1.0em}{--} \\
  \lstick{\ket{0}_1} &\qw & \gate{R_x(\theta_{1,0})} & \gate{R_y(\theta_{1,1})} & \gate{R_z(\theta_{1,2})} &\qw &\qw &\qw &\gate{R_x(\lambda_{1}d_1)} &\qw &\qw &\qw &\ctrl{-1} &\ctrl{1} &\qw &\qw &\qw &\qw &\qw &\ghost{\mathbf{x}_1(\psi)} &\qw & \meter\\
  \lstick{\ket{0}_2} &\qw & \gate{R_x(\theta_{2,0})} & \gate{R_y(\theta_{2,1})} & \gate{R_z(\theta_{2,2})} &\qw &\qw &\qw &\gate{R_x(\lambda_{2}d_2)} &\qw &\qw &\qw &\qw &\ctrl{-1} &\ctrl{1} &\qw &\qw &\qw &\qw &\ghost{\mathbf{x}_1(\psi)} &\qw & \meter\\
  \lstick{\ket{0}_3} &\qw & \gate{R_x(\theta_{3,0})} & \gate{R_y(\theta_{3,1})} & \gate{R_z(\theta_{3,2})} &\qw &\qw &\qw &\gate{R_x(\lambda_{3}d_3)} &\qw &\qw &\qw &\qw &\qw &\ctrl{-1} &\ctrl{-1} &\qw &\qw &\qw &\ghost{\mathbf{x}_1(\psi)} &\qw & \meter\\
  }} 
  \caption{ 
    \textbf{Illustration of a simple 4-qubit PQC architecture in the search space of EQAS-PQC.} This architecture, of which the genome encoding is $1-2-3-0$, is composed of 4 operations: 1) Variational PQC ($\mathbf{x}_1$) performs rotations on each qubits according to parameters $\theta$; 2) Data-encoding PQC ($\mathbf{x}_2$) performs rotations on each qubit according to the input data $d$ and scaling parameter $\lambda$; 3) Entanglement ($\mathbf{x}_3$) performs circular entanglement to all the qubits; 4) Measurement ($\mathbf{x}_0$) adds another Variational PQC ($\mathbf{x}_1$) and perform measurement to obtain the observable values.
  }
  \label{fig:pqc}
\end{figure*}

Given a fixed $n$-qubit system, a parametrized quantum circuit (PQC) is defined
by a unitary operation $U(s, \theta)$ that acts on the current quantum states
$s$ considering the trainable parameters $\theta$. In this work, we mainly
consider two types of PQCs: variational PQCs
(V-PQCs)~\citep{kandala2017hardware,benedetti2019parameterized} and
data-encoding PQCs (D-PQCs)~\citep{schuld2021effect,perez2020data}. The V-PQCs
are composed of single-qubit rotations $R_x$, $R_y$, $R_z$ with the rotation
angles as trainable parameters. The D-PQCs have a similar structure with
rotations, but the angles are the input data $d$ scaled by a trainable parameter
$\lambda$. The structures of both PQCs are depicted in Fig.~\ref{fig:pqc}, which
we describe in details later in Sec.~\ref{sec:encode}.

A recent work~\citep{jerbi2021parametrized} proposes to use an
alternating-layered architecture~\citep{schuld2021effect,perez2020data} to
implement parameterized quantum policies for RL, which basically applies an
alternation of V-PQC (followed by an entanglement) and D-PQC till the target
depth. While this architecture is simple and effective, it is obvious to see
that this general design can be easily modified and probably improved by
changing the placement of its components. In this work, we aim to optimize the
design of such PQC-based systems with architecture search methods.

\subsection{Quantum Architecture Search}

Early research~\citep{spector2004automatic} has shown the usage of genetic
programming to solve specific quantum computing problems from an evolutionary
perspective. Prior
works~\citep{tang2021qubit,las2016genetic,du2020quantum,lu2021markovian,niu2019universal,kuo2021quantum,zhang2020differentiable}
have explored the usage of common architecture search approaches in various
quantum computing applications such as quantum control, variational quantum
eigensolver, and quantum error mitigation. 
However, most of these works target on specific quantum computing problems and
try to optimize the quantum circuits in a hardware-efficient manner. 

More recently, a few approaches have been proposed to optimize the architectures
involving parameterized quantum circuits. \citet{grimsley2019adaptive} proposed
a method that iteratively adds parameterized gates and re-optimizes the circuit
using gradient descent. \citet{ostaszewski2021structure} proposed an
energy-based searching method for optimizing both the structure and parameters
for single-qubit gates and demonstrated its performance on a variational quantum
eigensolver. In this work, we take one step further and propose a more general
architecture search framework for hybird quantum-classical systems with both
parameterized and non-parameterized quantum operators, which aim to solve the
challenging learning problems such as RL.

\section{Method}

We propose EQAS-PQC, an Evolutionary Quantum Architecture Search framework for
constructing quantum learning models based on some fundamental PQC circuits.
While the proposed framework can be generally applied to various learning
problems, in this work, we choose to target on the challenging RL problems in
order to better illustrate the benefit of our method. In this section, we
describe the major components of EQAS-PQC including encoding scheme and search
process in detail.

\subsection{Encoding and Search Space}  \label{sec:encode}

Biologically inspired methods such as genetic algorithms (GAs) have been
successfully used in many search and optimization problems for decades. In most
cases, GAs refer to a class of population-based computational paradigms, which
simulate the natural evolution process to evolve programs by using genetic
operations (\eg, crossover and mutation) to optimize some pre-defined fitness or
objective functions. From this perspective, we may view the architectures of
quantum circuits as \textit{phenotypes}. Since the genetic operations usually
work with \textit{genotypes}, which are representations where the genetic
operations can be easily applied, we need to define an encoding scheme as the
interface for abstracting the architectures to genomes, where the genes are
different quantum operations.

The existing architectures of parameterized quantum
policies~\citep{jerbi2021parametrized} can be viewed as a composition of
functional quantum circuits that specify some computational schemes on a single
qubit or multiple qubits. In EQAS-PQC, we define four basic operation encodings
$\mathbf{x} = \{\mathbf{x}_0, \mathbf{x}_1, \mathbf{x}_2, \mathbf{x}_3\}$, and
the corresponding genes are represented as integers $\{0,1,2,3\}$. An
illustration of a simple PQC architecture in the search space of EQAS-PQC is
depicted in Fig.~\ref{fig:pqc}. More specifically, given a fixed $n$-qubit
state, we define the following operations:

\begin{itemize}
  \item $\mathbf{x}_1$: Variational PQC - A circuit with single-qubit rotations
  $R_x$, $R_y$, $R_z$ performed on each qubit, with the rotation angles as
  trainable parameters.
  \item $\mathbf{x}_2$: Data-encoding PQC - A circuit with single-qubit
  rotations $R_x$ performed on each qubit, with the rotation angles is the input
  scaled by trainable parameters.
  \item $\mathbf{x}_3$: Entanglement - A circuit that performs circular
  entanglement to all the qubits by applying one or multiple controlled-Z gates.
  \item $\mathbf{x}_0$: Measurement - A Variational PQC followed by measurement.
\end{itemize}

The outputs are computed by weighting the observables by another set of
trainable parameters for each output, with optional activation functions such as
\textsc{Softmax}. The architecture encoding/decoding is terminated when
approaching to $\mathbf{x}_0$.

It is easy to see that the search space of EQAS-PQC is dependent on the maximal
length of the genomes. Since the encoding will terminate when approaching to
$\mathbf{x}_0$, there will be cases where the same architecture is decoded from
different genomes. So the search space is the sum of possible operations (except
$\mathbf{x}_0$) for all the possible length less than the maximum length. In
other words, given a maximum length of the genomes $n$, the search space of
EQAS-PQC is
\begin{equation}
  \Omega_{\mathbf{x},n}=\sum_{i=1}^{n}(|\mathbf{x}|-1)^{i-1}
\end{equation}


\begin{table*}
  \caption{\textbf{RL environment specifications and hyperpameters.} (*: The reward
  function of MountainCar-v0 has been modified from the standard version as in
  OpenAI Gym, following the practices in \citet{jerbi2021parametrized,
  duan2016benchmarking}.)}
  \label{tab:gym}
  \centering
  \begin{tabular}{cccccc}
    \toprule
    \cmidrule(r){1-2}
    Environment & \# States/Qubits & Reward & Learning rates & $\gamma$ & Observables \\
    \midrule
    CartPole-v1 & $4$ & $+1$ & $0.01, 0.1, 0.1$ & $1.0$ & [$Z_0Z_1Z_2Z_3$]\\
    MountainCar-v0 & $2$ & $-1+height^*$ & $0.01,0.1,0.01$ & $1.0$ & [$Z_0,Z_0Z_1,Z_1$]\\
    \bottomrule
  \end{tabular}
\end{table*}

\begin{figure*}
  \centering
  \includegraphics[width=.49\linewidth]{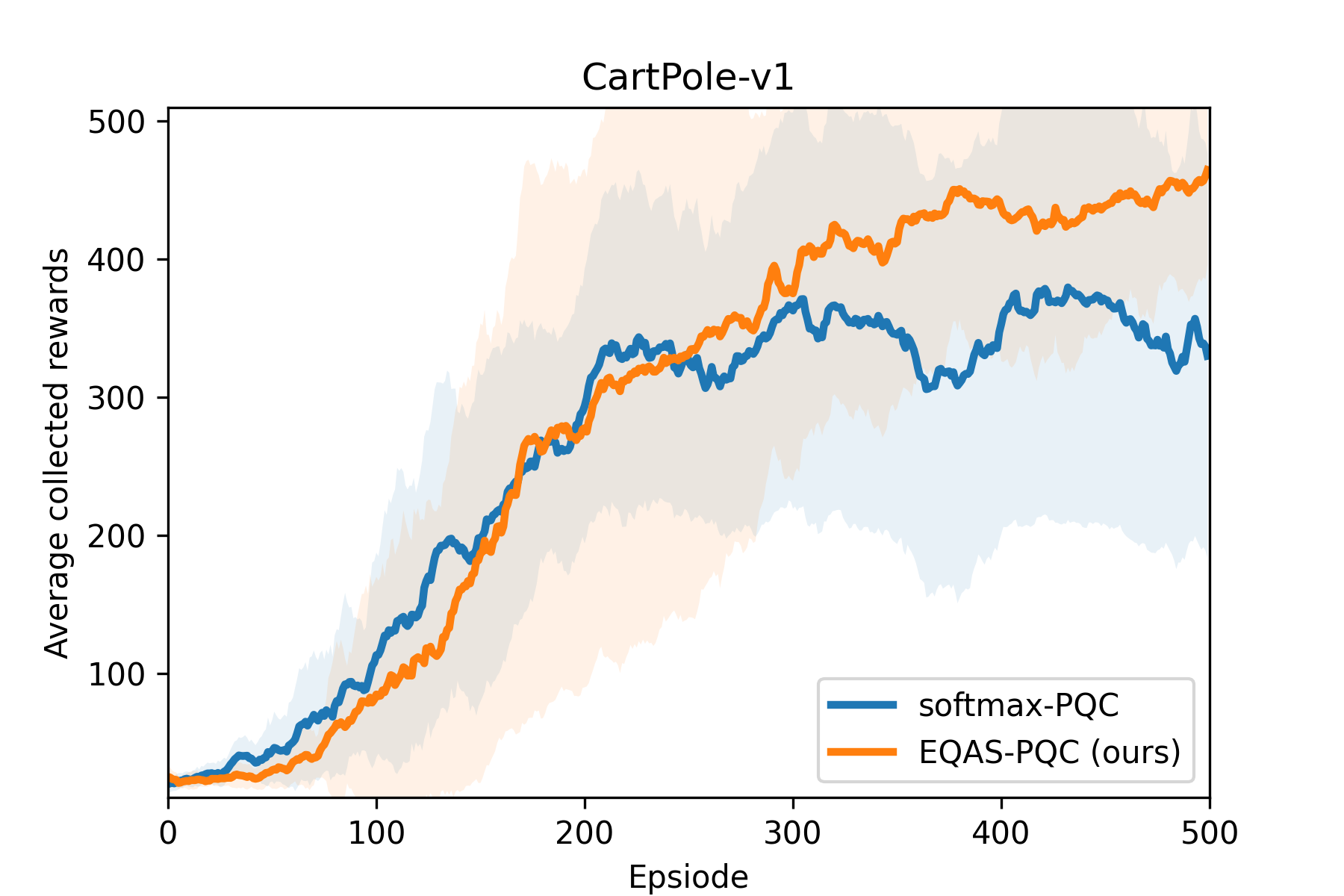}
  \includegraphics[width=.49\linewidth]{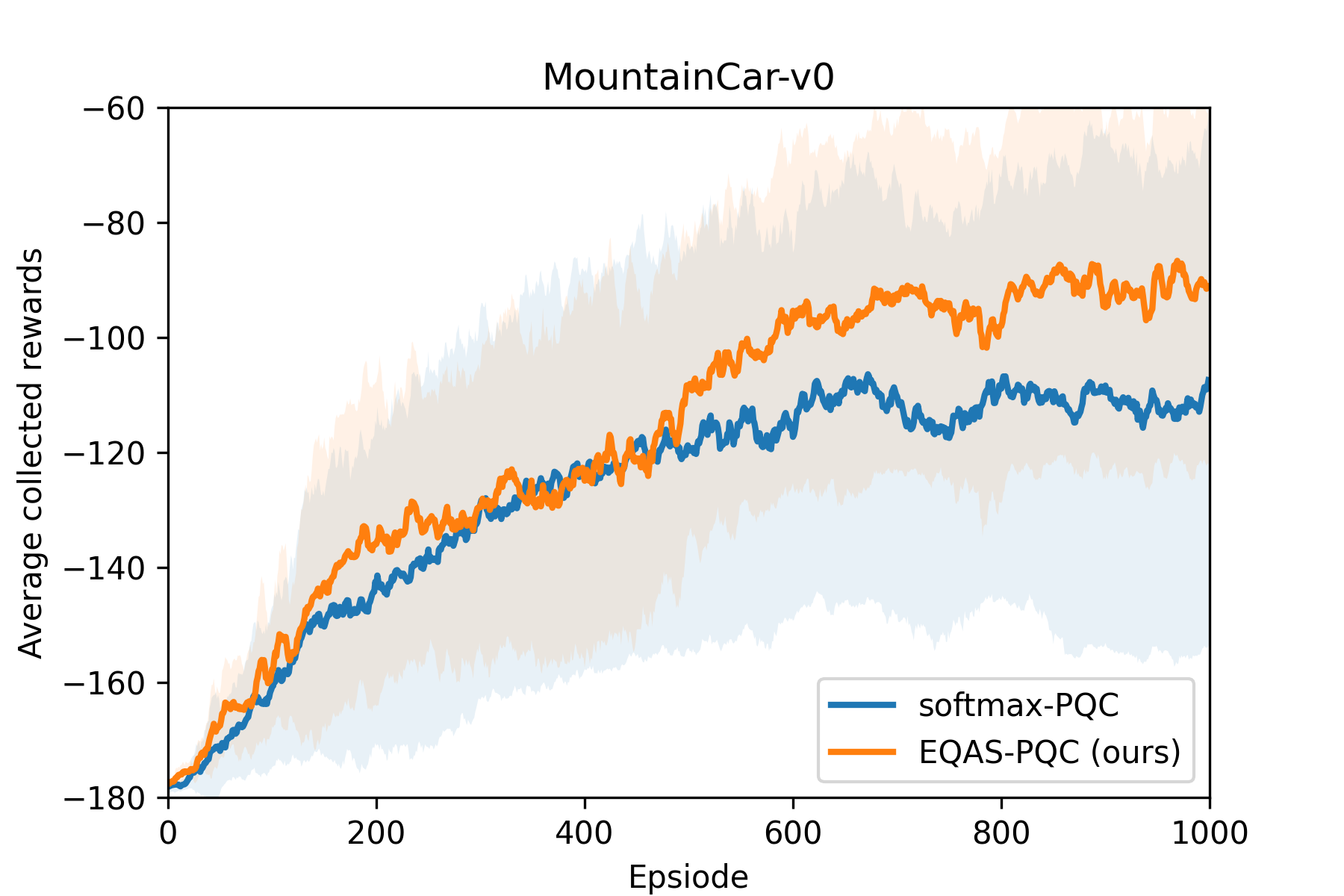}
  \caption{\textbf{Learning performance of EQAS-PQC on benchmark RL
  environments.} We plot the average learning curves (smoothed by a temporal
  window of 10 episodes)) over 10 randomly-initialized EQAS-PQC agents and
  \textsc{Softmax}-PQC agents in two benchmark RL environments (CartPole-v1 and
  MountainCar-v0) from OpenAI Gym. The shaded areas represent the standard
  deviation of the average collected reward.}
  \label{fig:rl}
\end{figure*}

\subsection{Search Process}

Similar to many other genetic algorithms, EQAS-PQC iteratively generates a
population of candidates (architectures) through genetic operations on the given
parents, and selects parents for the next generation based on fitness
evaluation. In this work, we adopt the Non-Dominated Sorting Genetic Algorithm
II (NSGA-II)~\citep{deb2000fast} to optimize the search process, with the
average collected rewards as the objective. NSGA-II has been successfully
employed in various single- and multi-objective optimization problems including
NAS~\citep{lu2019nsga}. 

The goal of EQAS-PQC is to discover diverse sequential combinations of quantum
operators and optimize the process with respect to the objective. Towards this
goal, we elaborate on the following components of the search process:

\subsubsection*{Crossover.} \hspace{1em} We use the two-point crossover to
perform recombination of the parent architectures to generate new offspring.
This method randomly choose two crossover points and swap the bits between the
parents, and has been widely used in search problems such as NAS. The intuition
is that sequential architectures for learning models usually require different
substructure for the beginning (input), middle (modeling), and ending (output)
of the architecture. From this perspective, the two-point crossover can
hopefully separate the three parts of the model and improve the architecture
through recombination.

\subsubsection*{Mutation.} \hspace{1em} To enhance the diversity of architectures in the
population, we also add polynomial mutation, which has been widely used in
evolutionary optimization algorithms as a variation operator. Given the specific
encoding, the mutation is operated in integer space, and will allow the search
to potentially reach any possible genomes in the search space.

\subsubsection*{Duplicate elimination.} \hspace{1em} It is worth noting that, given the
proposed encoding scheme, some different genomes may be decoded to the same
architecture, \eg, any operations after $\mathbf{x}_0$ does not change the
architecture. To maintain the diversity of population, we additionally eliminate
those duplicate architectures with different genomes.

\subsubsection*{Fitness evaluation.} \hspace{1em} For each generation, we decode the
population to different architectures, and use the architectures to construct
\textsc{Softmax}-PQC~\citep{jerbi2021parametrized} policies for the RL agents.
While EQAS-PQC can be easily extended to optimize for multiple objectives, in
this work, we demonstrate by using the learning performance of the RL agents as
a single objective for fitness evaluation. The learning performance is computed
as the average episode reward to represent the area under the learning curve.

\section{Experiments}

In this section, we describe the experimental setup and implementation details
of EQAS-PQC on the classical benchmark RL environments. We also show the
empirical results of our method compared to the prior work (\textsc{Softmax}-PQC
by \citet{jerbi2021parametrized}) to demonstrate the advantage of using EQAS-PQC
against commonly-used alternating-layer PQC architectures.

\subsection{RL Environments}

In this work, we consider two classical RL benchmark environments from the
OpenAI Gym~\citep{brockman2016openai}: CartPole and MountainCar. Both
environments have continuous state spaces and discrete action spaces, and have
been widely used in RL research, including prior works on quantum
RL~\citep{jerbi2021parametrized,skolik2021quantum}. The CartPole task is to
prevent the pole from falling over by controlling the cart. For MountainCar, the
goal is to drive up the mountain by driving back and forth to build up momentum.
More detailed description can be found in \citet{brockman2016openai}. The
specifications are presented in Table~\ref{tab:gym}, where the reward is the
step reward and $\gamma$ is the discount factor for future rewards.

\subsection{Implementation Details}

\subsubsection*{Search algorithm.} \hspace{1em} For both environments,
EQAS-PQC uses a population size of 20 and runs for 20 generations. The maximum
length of architecture is set to 30. We also reduce the total number of episode
to a factor of 0.8 in the search process to improve the efficiency of evolution.
The main search framework is implemented using the
\textit{pymoo}~\citep{blank2020pymoo}.

\subsubsection*{RL training during search.} \hspace{1em} For each generated
PQC policy, we learn the policy for a single trial, and calculate the average
episode rewards as its learning performance. We set the hyperparameters such as
learning rates and observables following the general practice in
\citet{jerbi2021parametrized}, which are also summarized in Table~\ref{tab:gym}.
All the agents are trained using REINFORCE~\citep{williams1992simple}, which is
a basic Monte Carlo policy gradient algorithm. We additionally apply the
value-function baseline~\citep{greensmith2004variance} in MountainCar to
stabilize the Monte Carlo process, which has been commonly used in recent RL
methods~\citep{duan2016benchmarking}. The quantum circuits are implemented using
Cirq~\citep{hancock2019cirq} and the learning process is simulated using
TensorFlow~\citep{abadi2016tensorflow} and TensorFlow
Quantum~\citep{broughton2020tensorflow}.

\subsubsection*{Performance evaluation.} \hspace{1em} For the final results,
we take the best performing architecture for each environment and evaluate it
for 10 trials (500 episodes for CartPole and 1000 episodes for MountainCar). To
compare with prior work, we also evaluate the alternating-layer architecture
(\textsc{Softmax}-PQC) as used in \citet{jerbi2021parametrized}, which can be
viewed as a special case in the search space of EQAS-PQC.

\subsection{Results}

We evaluate the general performance of the proposed EQAS-PQC and the experiment
results are presented as follows. There are two goals for our experiment: 1) to
show that our method is able to find PQC architectures with better learning
performance as well as a similar computation cost to prior work; 2) to discover
the performance-critical design choices of PQCs in addition to the commonly-used
alternating-layer architecture. To illustrate the above points, we first apply
EQAS-PQC to two classical RL benchmark environments and obtained the best
performing architectures, and then conduct two analysis on the resulting
architectures.

\subsubsection*{Learning Performance}\hspace{1em}
We evaluate and visualize the average learning performance over 10 trials of the
best-performing architecture searched by EQAS-PQC and the one used in
\textsc{Softmax}-PQC~\citep{jerbi2021parametrized}, as shown in
Fig.~\ref{fig:rl}. The corresponding genomes of the EQAS-PQC architectures for
the two RL environments are:
\begin{itemize}
  \item CartPole: \\
  $3-3-2-3-3-1-2-1-3-2-3-2-0$
  \item MountainCar: \\
  $3-1-2-3-1-2-2-2-3-2-1-1-1-3-2-0$
\end{itemize}
To ensure a fair comparison, for \textsc{Softmax}-PQC, we use the depth of 6,
resulting in an architecture with length 19, which is larger than the resulting
architectures searched by EQAS-PQC for both environments. Thus, we can conclude
that our method is able to find PQC architectures that significantly outperform
the standard alternating-layer PQC.

\begin{figure}
  \centering
  \includegraphics[width=.99\linewidth]{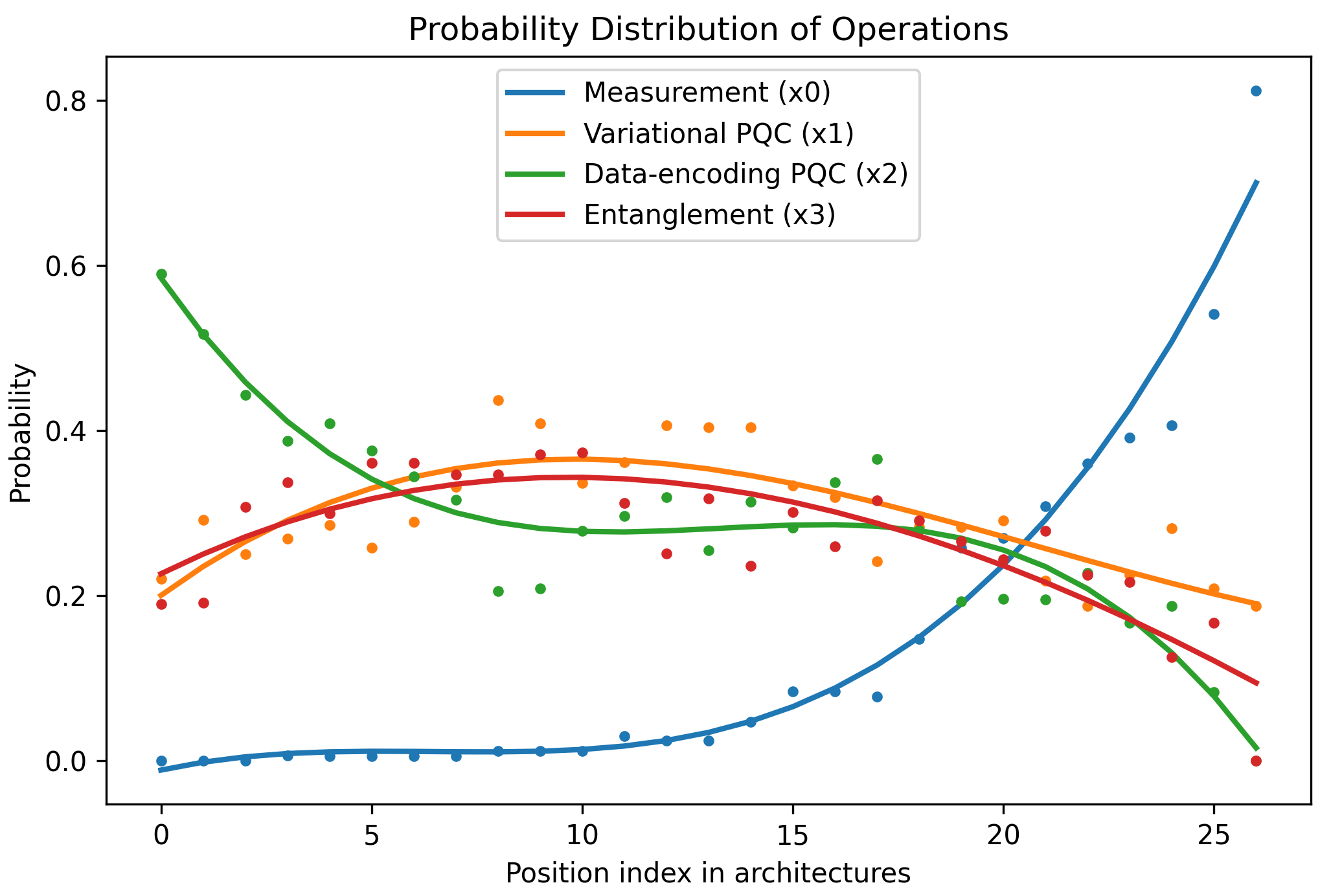}
  \caption{Probability distribution of quantum operations in top-performing PQC
  architectures. We select 20 top-performing architectures searched by EQAS-PQC
  (10 for each RL environment), and calculate the probability distributions of
  operations at each position in the architecture.}
  \label{fig:prob}
\end{figure}

\subsubsection*{Probability Distribution of Quantum Operations}\hspace{1em} We
also want to tell the reason why architectures found by EQAS-PQC is able to have
better performance. To illustrate this, we calculate the probability
distribution of all the encoded operations at each position in the architecture,
and visualize in Fig.~\ref{fig:prob}. The probabilities are smoothed by
a window of length 5 and the fitted lines are polynomials.

From the plot, we can first see that the Variational PQC has a similar frequency
as entanglement, which aligns with the design of alternating-layer PQC. However,
the frequency of Data-encoding PQC has an obvious decreasing trend, indicating
that it is better to have more Data-encoding PQCs at the beginning of the
architecture. This finding is intuitive and can be referred to the general
machine learning modeling practices, where data input is usually at the
beginning of the modeling. Finally, the probability of Measurement does not
increase till the end of architecture, meaning that most of the optimal
architectures have a similar length of around 20. This also shows the advantage
of PQCs that shallow architectures with a small number of qubits are able to
handle the challenging RL problems, which has also been proved in prior
works~\citep{jerbi2021parametrized}.

\section{Conclusion and Future Work}

In this work, we propose EQAS-PQC, an evolutionary quantum architecture search
framework for parameterized quantum circuits. EQAS-PQC uses the population-based
genetic algorithm to evolve PQC architectures by exploring the search space of
quantum operations. Experimental results show that our method can significantly
improve the performance of hybrid quantum-classical systems in solving benchmark
reinforcement learning problems. In addition, we also analyze the probability
distributions of quantum operations in top-performing architectures, and
identifies design choices that are essential to the performance of PQC-based
architectures.

One limitation to our work is that the experiments are conducted using a
simulation backend for quantum circuits. For future work, we expect to extend
our work to use real quantum computers and add more objectives to the
consideration of evolutionary, such as quantum noise and hardware efficiency.

\begin{acks}
  This work is supported by the National Science Foundation under Grant No.
  1617087. Any opinions, findings, and conclusions expressed in this publication
  are those of the authors and do not necessarily reflect the views of the
  National Science Foundation. The authors would like to thank Sofiene Jerbi for
  providing implementation details for reproducing baseline results. The
  authors would like to thank Edward Pantridge, Thomas Helmuth, Ryan Boldi,
  and Anil Saini for their valuable comments and helpful suggestions.
  
\end{acks}


\bibliographystyle{ACM-Reference-Format}
\bibliography{../quantum}


\end{document}